%% file: gendis-ijcai19.tex
\documentclass{article}
\usepackage{ijcai19}

\usepackage{times}
\usepackage{soul}
\usepackage{url}
\usepackage[hidelinks]{hyperref}
\usepackage[utf8]{inputenc}
\usepackage[small]{caption}
\usepackage{graphicx}
\usepackage{amsmath}
\usepackage{booktabs}
\usepackage{algorithm}
\usepackage{algorithmic}
\usepackage{hyperref}       %
\usepackage{url}            %
\usepackage{booktabs}       %
\usepackage{amsfonts}       %
\usepackage{nicefrac}       %
\usepackage{microtype}      %
\usepackage{amsmath,amssymb}
\usepackage{lastpage,fancyhdr,graphicx}
\usepackage{rotating}
\usepackage{xr}
\usepackage{csvsimple}

\newcommand{\vecx}{\mathbf{x}}
\newcommand{\vecy}{\mathbf{y}}

\newcommand{\realnum}{\mathcal{R}}
\newcommand{\gaussd}{\mathcal{N}}

\newcommand{\dataset}{\mathcal{D}}
\newcommand{\lowerb}{\mathcal{L}}
\newcommand{\expect}{\mathcal{E}}
\newcommand{\KL}{\mathrm{KL}}

\title{Group-based Learning of Disentangled Representations \\
with Generalizability for Novel Contents}

\author{ Haruo Hosoya \affiliations ATR International, Kyoto, Japan \emails hosoya@atr.jp } 

\begin{document}
\maketitle

\begin{abstract}
Sensory data are often comprised of independent content and transformation factors.  For example, face images may have shapes as content and poses as transformation.  To infer separately these factors from given data, various ``disentangling'' models have been proposed.  However, many of these are supervised or semi-supervised, either requiring attribute labels that are often unavailable or disallowing for generalization over new contents.  In this study, we introduce a novel deep generative model, called group-based variational autoencoders.  In this, we assume no explicit labels, but a weaker form of structure that groups together data instances having the same content but transformed differently; we thereby separately estimate a group-common factor as content and an instance-specific factor as transformation.  This approach allows for learning to represent a general continuous space of contents, which can accommodate unseen contents.  Despite the simplicity, our model succeeded in learning, from five datasets, content representations that are highly separate from the transformation representation and generalizable to data with novel contents.  We further provide detailed analysis of the latent content code and show insight into how our model obtains the notable transformation invariance and content generalizability.  
\end{abstract}

\section{Introduction}
\label{sec:intro}

Sensory data are often composed of multiple independent factors.  For example, 3D face images usually consist of the content, representing how the face is shaped, and the transformation, representing how the face is posed or expressing.  Humans are generally good at inferring and separating such factors from given inputs without much supervision.  Moreover, once discovering such factors, one can automatically generalize this knowledge for novel contents, e.g., faces of unseen identities.  How can such ability be achieved computationally?  

The content-transformation separation problem \cite{Tenenbaum:2000wj} has recently been attracting much attention, in accordance with the progress of deep generative techniques such as variational autoencoders (VAE) \cite{Kingma:2013tz} and generative adversarial learning (GAN) \cite{Goodfellow:2014td}, and nowadays often called ``disentangling.''   For this purpose, a number of supervised or semi-supervised models have been proposed exploiting labels in various ways.  The most typical approach learns a generative model by explicitly supplying class labels to the content variable while extracting the remaining factor in the transformation variable \cite{Kingma:2014uq,Cheung:2014ug,Siddharth:2017ud}.  However, since the content representation in such approach is often restricted to the classes seen during training, it does not generalize for unseen classes.  Other approaches use a more sophisticated method, such as adversarial learning, that exploits labels so as to make the content representation as irrelevant as possible to transformation \cite{Wang:2017taa,Lample:2017vt,Mathieu:2016tn}.  Although these approaches can potentially allow for generalization over new contents, their requirements of specific kinds of label are often difficult to fulfill, e.g., attribute labels corresponding to transformation.  

\begin{figure*}[h]
\begin{center}
\includegraphics[width=14cm]{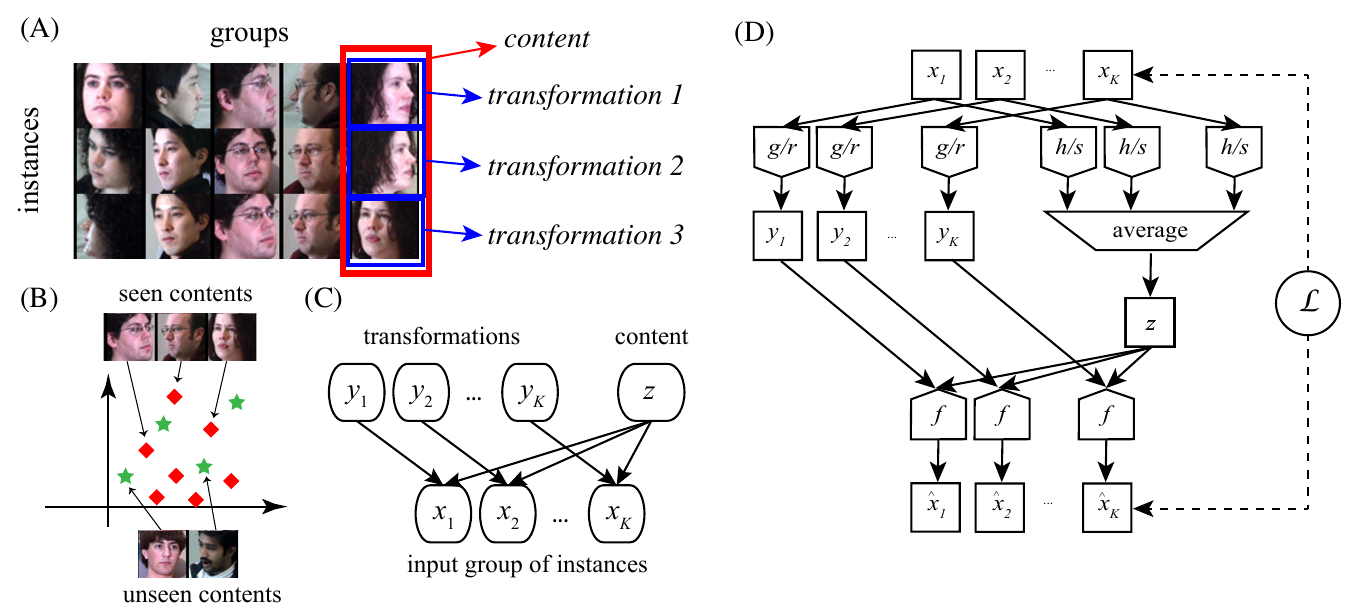}
\caption{(A) Example of grouped data.  Each column is a group of face images of the same person (content) with different views (transformation).  (B) A continuous space of contents learned from training data can accommodate novel contents.  (C) The graphical model.  Each instance $x_k$ is generated from the instance-specific transformation variable $y_k$ and the group-common content variable $z$.  (D) Algorithmic outline of GVAE learning.  The encoders $g$ and $r$ given an individual image $x_k$ compute the corresponding transformation representation $y_k$, while the encoders $h$ and $s$ followed by averaging compute the common content representation $z$.  The decoder $f$ then reconstructs each image $\hat{x}_k$ by combining the corresponding transformation $y_k$ and the common content $z$.  (See text for details.)}
\label{fig:model}
\end{center}
\end{figure*}

In this study, we propose a new deep generative model, called group-based variational autoencoders (GVAE), for content-transformation separation.  In GVAE, we do not require explicit labels in the dataset.  Instead, we assume a weaker structure that groups together multiple data instances (like images) containing the same content but transformed differently.  Then, our model learns to extract the content as the factor common within each group and the transformation as the factor specific to each instance.  For example, from the set of groups of face images as in Figure~\ref{fig:model}A, the facial shape can be extracted as the group-common factor and the facial pose and expression as the instance-specific factor.  Our grouping assumption is quite general in the sense that such structure can be derived not only from class-labeled data by grouping instances of the same class, but also from video data by grouping frames in the same or nearby sequence.   Further, this approach allows for generalization over novel contents: we use a continuous variable to form a general ``space'' of content (e.g., a space of facial shapes) and thereby can represent an infinite number of contents including those not seen during training (Figure~\ref{fig:model}B).

Our specific contributions are (i) to describe a novel learning method that takes a grouped dataset (with no other labels) and separately extracts group-common and instance-specific factors, (ii) to show, qualitatively and quantitatively, that this method learns highly separate representation of content and transformation and generalizes it to new contents, for five datasets, and (iii) to provide detailed analysis of latent code to reveal insight into the generalizable disentangling.

\section{Related Work}
\label{sec:related}

As already mentioned, a number of supervised or semi-supervised approaches have been developed for disentangling.  The simplest approach is to explicitly supply the content information and infer the transformation variable as the remaining factor \cite{Kingma:2014uq,Cheung:2014ug,Siddharth:2017ud}.  However, the prior studies in this approach have typically supplied class labels for supervision (since these are the only labels available in most datasets), represented them in categorical variables, and estimated a generative model conditioned on the class.  Since such estimated model would have little useful information for unseen classes, generalization for new contents is difficult.

In other supervised approaches, more sophisticated mechanisms have been devised to exploit labels for discovering a transformation-invariant content representation in a continuous variable.  For example, \cite{Wang:2017taa} has introduced a mapping (deep net) from labels to latent code while maintaining consistency with another mapping from input images to the same latent code; \cite{Lample:2017vt,Mathieu:2016tn} have used adversarial learning to make the content code as irrelevant as possible to transformation using labels.  While effective disentangling can thus be achieved with the aid of labels, requirement for labels, in particular, attribute labels corresponding to transformation, seems rather restrictive.

Concurrently with ours, one recent study has developed another group-based method called Multi-Level VAE (MLVAE) \cite{Bouchacourt:2018vt}.  In this, they adopt a sophisticated technique called ``evidence accumulation'' for estimating the group-common factor.  However, as we show later, this particular technique has an unfortunate property that the learned content representation often becomes dependent on the transformation, which potentially conflicts the goal of disentangling (Section~\ref{sec:coding}).  Indeed, due to this, MLVAE often results in quantitatively poorer disentangling representations compared to our GVAE (Section~\ref{sec:results}).  In \cite{Chen:2017tk}, a group-based GAN method has been proposed for learning separate representation of content and transformation.  However, their goal is rather different from ours, as they focus on randomly generating new realistic images while lacking a way to  infer content and transformation information from given images.

Some studies have investigated disentangling for sequential data like video \cite{Yang:2016wg,Denton:2017uf}.  They crucially assumed that the content variable remains similar in consecutive frames, as inspired by the temporal coherence principle \cite{Foldiak:1991up}.  Although this idea is related to the group-based approach, the sequence-based methods have used additional mechanisms to exploit ordering among data instances, e.g., a recurrent neural network to predict future states from past states \cite{Yang:2016wg} or adversarial training to take temporal structure into account \cite{Denton:2017uf}.

Some unsupervised approaches require no label or grouping in data, but optimizes the efficiency of the latent representation, either by maximizing mutual information between hidden variables in a GAN-based model \cite{Chen:2016tp} or by adjusting a so-called $\beta$-term in a variational lower bound \cite{Higgins:2016vm}.  Neither study reported generalizability over novel contents.

\section{Methods}
\label{sec:method}

\subsection{Model}
\label{sec:model}

We assume a dataset $\dataset$ consisting of $N$ groups, in which each group has $K$ data instances, $(x_1^{(n)},\ldots,x_K^{(n)})$, where $x_k^{(n)}\in \realnum^D$ is a data instance indexed by the group number $n$ and the instance number $k$.  For example, Figure~\ref{fig:model}A shows a set of 5 groups of 3 data instances, where each instance is an image.  (For brevity, we sometimes elide the superscript $(n)$ below.)  We assume independence between groups but not instances within a group.  We intend that each group contains the same content with (possibly) different transformations.  In other words, the factor common among the instances correspond to the content, while the factor differentiating them correspond to the transformation.  (We do not require alignment in transformation at each instance number.  Also, we do not forbid that different groups happen to contain the same content.)  

For such grouped data, we consider the following probabilistic generative model with two types of hidden variables: the (instance-specific) transformation variables $y_1,\ldots,y_K\in\realnum^L$ and the (group-common) content variable $z\in\realnum^M$ (Figure~\ref{fig:model}C):
\begin{align}
	p(z) &= \gaussd(0,I) \\
	p(y_k) &= \gaussd(0,I) \\
	p_\theta(x_k|y_k,z) &= \gaussd( f_\theta(y_k,z), I )
\end{align}
for $k=1,\ldots,K$.  Here, $f_\theta$ is a decoder deep net with weight parameters $\theta$.  
In the model, the content $z$ or each transformation $y_k$ is first generated from the standard Gaussian prior.  Then, each instance $x_k$ is generated from the decoder $f_\theta$ applied to the corresponding transformation $y_k$ and the common content $z$, added with Gaussian noise of unit variance.

\subsection{GVAE}
\label{sec:gvae}

For learning, we extend the VAE approach introduced by \cite{Kingma:2013tz}, which uses encoder models based on deep nets to approximate posterior distributions for hidden variables.  (Figure~\ref{fig:model}D illustrates the outline of the learning algorithm.)  First, we estimate each transformation $y_k$ from the corresponding input instance $x_k$ as follows:
\begin{align}
	q_{\phi,\xi}(y_k|x_k) &= \gaussd\left( g_\phi(x_k), r_\xi(x_k)\right) 
\end{align}
where we use an encoder deep net $g_\phi$ with weight parameters $\phi$ for estimating the mean and another deep net $r_\xi$ (positively valued) with weight parameters $\xi$ for estimating the dimension-wise variances.\footnote{Since we consider only Gaussians with diagonal covariances, we specify a vector of variances in the second parameter to Gaussian distribution as convention.}   For inference of content, we could likewise assume a pair of deep nets to estimate the content $z$ from all instances $x_1,\ldots,x_K$, but it cannot exploit symmetry in the instances in the same group.  Instead, we take the following simpler approach:
\begin{align}
	q_{\psi,\pi}(z|x_1,\ldots,x_K) &= 
	\gaussd\left( \frac{1}{K} \sum_{k=1}^K h_\psi(x_k), \frac{1}{K} \sum_{k=1}^K s_\pi(x_k) \right) \label{eq:gvae-inf} 
\end{align}
That is, the encoder deep nets $h_\psi$ and $s_\pi$ first each estimate the mean and variance of the posterior distribution of the individual content for each instance $x_k$.  Then, we infer the common content $z$ for all instances as the average of all the individual contents.  Note that, therefore, the variance of $z$ becomes the average of the individual variances.  The intention here is that, as we attempt to reconstruct each instance $x_k$ by the common content $z$ with the individual transformation $y_k$, all the individual contents $h_\psi(x_1),\ldots,h_\psi(x_K)$ are encouraged to converge to an equal value in the course of learning.  Thus, $z$ will eventually become a common factor of all $x_k$, while each $y_k$ will become a differentiating factor. 

To train the model, we consider the following variational lower bound of the log likelihood:
\begin{align}
	\log p_\theta(\vecx) 
	&\geq \expect_{q_{\phi,\xi,\psi,\pi}(\vecy,z|\vecx)}\left[\sum_{k=1}^K \log p_\theta(x_k|y_k,z)\right] \nonumber \\
	&\quad - \sum_{k=1}^K \KL( q_{\phi,\xi}(y_k | x_k) \parallel p(y_k) ) \nonumber \\
	&\quad - \KL( q_{\psi,\pi}(z | \vecx) \parallel p(z) ) = \lowerb(\vecx) \label{eq:obj}
\end{align}
where $\vecx=(x_1,\ldots,x_K)$ and $\vecy=(y_1,\ldots,y_K)$.  Then, our goal is to maximize the lower bound for the whole dataset $\dataset$:
\begin{align}
	\lowerb(\dataset) 
	= \frac{1}{N}\sum_{n=1}^{N}\lowerb(\vecx^{(n)}) 
	\label{eq:obj-full}
\end{align}
This can be solved by the standard VAE techniques (stochastic variational Bayes, reparametrization, etc.; \cite{Kingma:2013tz}).  

\begin{figure*}[t]
\begin{center}
\includegraphics[width=16cm]{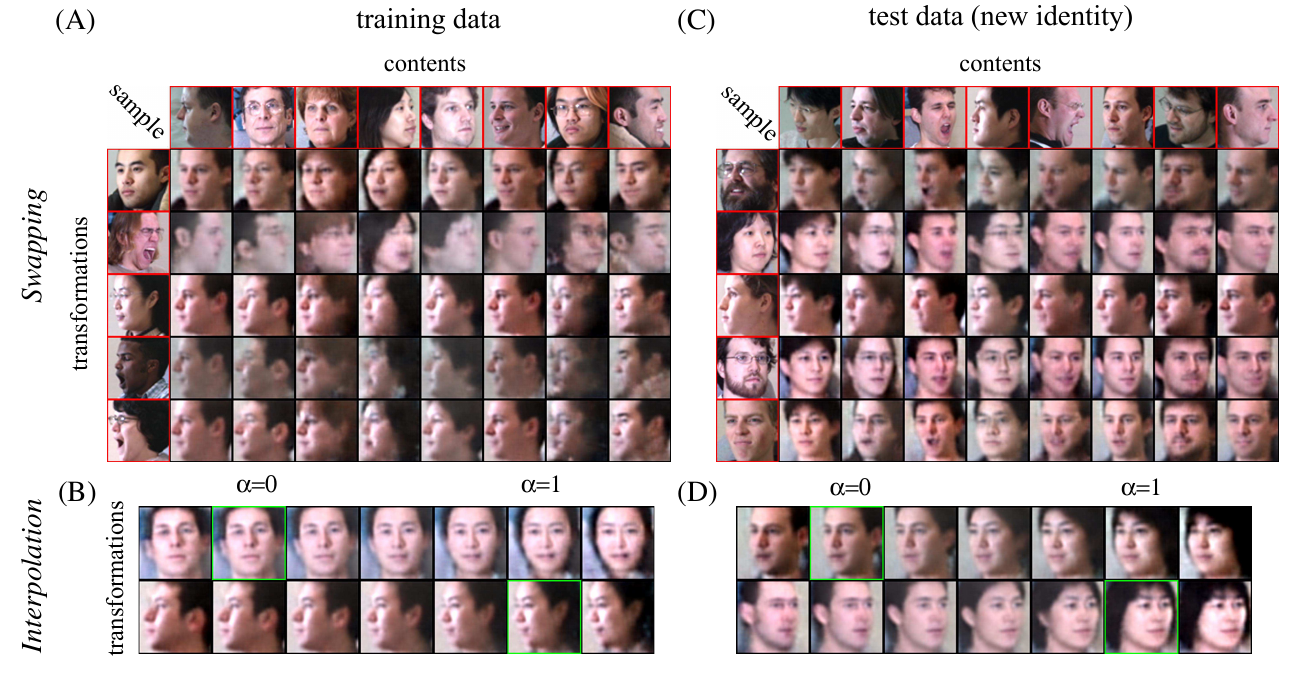}
\caption{Illustration of generalizable disentangling in a GVAE model trained with Multi-PIE.  (A) Swapping for training images.  Each image in the matrix was generated from the content representation of a sample image in the top row and the transformation representation of another sample image in the left-most column.  Red box: sample image.  %
(B) Interpolation for training images.  For two training sample images, each image was generated from a linear interpolation (or extrapolation) of the content representations of the two sample images, in conjunction with the transformation representation of either sample image.  Green box: images corresponding to the sample.  %
(C,D) Analogous swapping and interpolation for test images.  The test images have distinct facial identities from the training images.
}
\label{fig:result-multipie}
\end{center}
\end{figure*}

\subsection{Experimental Set-up}
\label{sec:exper}

We prepared the following five datasets.  %
(1) Multi-PIE: multi-viewed natural face images derived from \cite{Gross:2010fj}; grouping by the subject and (cloth/hair) fashion while varying the view and expression; the training and test sets with disjoint 268 and 69 subjects.  %
(2) Chairs: multi-viewed synthetic chair images derived from \cite{Dosovitskiy:2015ux}; grouping by the chairs type while varying the view; the training and test sets with disjoint 650 and 159 types.  %
(3) KTH: image frames from video clips of human (only pedestrian) motion derived from \cite{Schuldt:2004ee}; grouping by the video clip while varying the position of the subject; the training and test sets with disjoint 20 and 5 subjects.  %
(4) Sprites: multi-posed synthetic game character images \cite{Reed:2015uw}; grouping by the character id while varying the pose and wear; the training and test sets with disjoint 500 and 100 ids.  %
(5) NORB: multi-viewed toy images \cite{LeCun:2004if}; grouping by the toy id varying the view and lighting; the training and test sets with disjoint 25 and 25 ids.
Note that labels were used only to form groups for some datasets but never explicitly given to the training algorithm; no label was used to form groups in KTH (video data).
See Appendix~\ref{sec:dataset-add} for details of these datasets.

For each dataset, we built a number of GVAE models with convolutional neural nets for the encoders ($g$, $r$, $h$, and $s$) and a deconvolutional neural net as the decoder ($f$).
Each encoder had three convolution layers followed by two fully connected layers (intervened with RELU nonlinearity); each decoder had a reverse architecture.  The top layer of encoder $r$ or $s$ had nonlinearity to ensure positivity: $F(a)=\exp(a/2)$.  
We used the same architecture for all models with the transformation dimension $L=3$ (except $L=2$ for Chairs and KTH) and the content dimension $M=100$.  %
See Appendix~\ref{sec:architecture-add} for details of the architecture.  

We found it crucial to choose a very low dimension for the transformation variable ($L=2$ or $3$).  In this way, the transformation variable was endowed with a just enough space to encode the instance-specific factor and no spare space to encode the common factor.  If the dimensionality was instead much raised, then the transformation variable would learn to represent all aspects of inputs including the contents and thereby the content variable would become degenerate.  Apart from this, the results were generally stable across choices of other parameters such as deep net architectures.

To train each model, we first randomly initialized the weight parameters of the encoders and decoder and then optimized the objective~\eqref{eq:obj-full} with respect to the weight parameters using the training set.    %
Training proceeded by mini-batches (size 100), where each group was formed on the fly by randomly choosing 5 images according to the dataset-specific grouping policy described above ($K=5$).  %
We used Adam optimizer \cite{Kingma:2014us} with the recommended optimization parameters.  %
During training, we monitored effectiveness of each latent dimension (s.d. $\geq 0.1$) and discarded   models with very low effective dimensions (0 for $y$, or $\leq 3$ for $z$).

\begin{table*}[t]
\begin{center}
\begin{tabular}{|c|c|c|c|c|c|c|c|}
\hline%
success rate (\%) & \multicolumn{3}{c|}{1-shot classification} & \multicolumn{3}{c|}{5-shot classification} & chance \\
\hline\hline 
& GVAE & MLVAE & VAE & GVAE & MLVAE & VAE &  \\
\hline 
\input{figs/nshot.txt}
\hline 
\end{tabular}
\caption{Quantitative comparison of generalizability of disentangled representations evaluated by few-shot classification.  The success rate (mean and s.d.) is shown for each method and for each dataset in the 1-shot or 5-shot case, along with the chance level.}
\label{tbl:fewshot-comp}
\end{center}
\end{table*}

\begin{figure*}[t]
\begin{center}
\includegraphics[width=15cm]{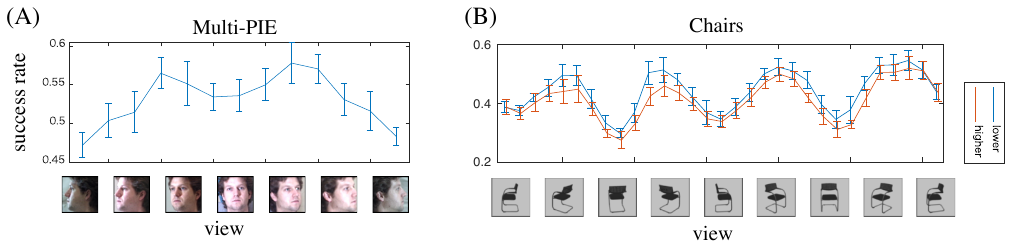}
\caption{(A) The success rate (y-axis) for each view (x-axis; two down-looking views are omitted) from a GVAE model trained with Multi-PIE.  (B) Analogous result for Chairs.  (Two colors correspond to different vertical angles of view; see legend).   
}
\label{fig:fewshot}
\end{center}
\end{figure*}

\section{Results}
\label{sec:results}

\subsection{Examples}
\label{sec:disentangling}

We first illustrate the learned disentangled representations in two ways.  {\em Swapping}: given two lists of inputs $x_1,\ldots,x_I$ and $x_1',\ldots,x_J'$, we show a matrix of images, each generated from the content representation estimated from an input $x_i$ and the transformation representation estimated from another input $x_j'$: $f(g(x_j'),h(x_i))$.  {\em Interpolation}: given two inputs $x_1$ and $x_2$, we show a matrix of images, each generated from a linear interpolation of the content representations estimated for both inputs, with the transformation representation for either input: $f(g(x_i),(1-\alpha) h(x_1)+\alpha h(x_2))$ for $0\leq \alpha\leq 1$ for $i=1,2$; extrapolation is also possible with $\alpha$ outside this range.  Note that we estimated the latent representation from a single image (with no grouping) and did not use the variance encoders $r$ and $s$.  

Figure~\ref{fig:result-multipie}A shows a swapping matrix for a GVAE model trained with Multi-PIE dataset.  The top row and the left-most column show two lists of sample images in the training set.  In the matrix, we can observe reasonably successful disentangling: each generated image reflects the subject of the corresponding input in the first list and the view of the corresponding input in the second list.  Figure~\ref{fig:result-multipie}B shows an interpolation matrix from the same model for sample training images.  Note the smooth transition of the generated images from one identity to another.  Also some generated images for extrapolated contents exhibit exaggerated facial features (e.g., hair).

Figure~\ref{fig:result-multipie}CD shows swapping and interpolation matrices for sample images in the test set.  Generalization of disentangling to the test case can be observed with a quality more or less similar to the training case.  This is quite remarkable, given the fact that none of the subjects here had been seen during the training.

\subsection{Quantitative Comparison}
\label{sec:fewshot}

We next show the results of our quantitative evaluation.  We here use accuracy of few-shot classification as criterion.  The rationale is that, since the learned content representation is expected to eliminate information on transformation, this should allow for transformation-invariant recognition of objects.  In particular, if the content representation is highly independent from transformation and highly generalizable for novel contents, then it should ideally be sufficient to hold the content representation of a single example of each unseen class in order to classify the remaining images (one-shot classification).  Thus, the better the generalizable disentangled representation is, the more accurate the few-shot classification should be.

Our evaluation procedure goes as follows.  For a dataset, we formed a split of the test images into the gallery including $S$ random images for each class and the probe including the remaining images ($S$-shot case).  Here, a class referred to each subject/fashion combination in Multi-PIE (378 test classes), to each chair type in Chairs (159 test classes), to each video clip id in KTH (240 test classes), to each sprite id in Sprites (100 test classes), and to each toy id in NORB (25 test classes).  Then, for a trained model, we classified each probe image as the class of the gallery image that had the maximal cosine similarity with the input probe image in the space of content variable $z$; classification succeeded if the inferred class matched the actual class; the success rate was averaged over 100 different splits.  %
We examined 10 separately trained models (different only in the initial weights) for each dataset.  

For comparison, we also trained models with the MLVAE method \cite{Bouchacourt:2018vt}, another group-based method; the training condition was the same.  In addition, we trained GVAE models with no grouping ($K=1$).  This in fact corresponds to a basic VAE model with a single variable since the content and transformation variables, without grouping, can be integrated without loss of generality.  

Table~\ref{tbl:fewshot-comp} summarizes the result, showing the success rate (mean and s.d. over the 10 model instances) of one-shot or five-shot classification for each method and for each dataset.  In most cases, GVAE outperformed the other methods.  In particular, for Multi-PIE and Sprites datasets, the scores for GVAE were much higher than MLVAE.  VAE models always performed poorly (though far better than the chance level); this condition in fact failed disentangling with no clear content-transformation separation and often corrupted generated images. Thus, grouping was indeed crucial.

As a side interest, we wondered which view of 3D objects led to more successful recognition.  Figure~\ref{fig:fewshot} shows the view-wise success rates of one-shot classification in GVAE models for Multi-PIE and Chairs.  We found that, in both cases, diagonal views always gave better success rates than profile or frontal views.  This result is intuitive since we can perceive better the entire shape of a face or chair from diagonal views than other views.  Perhaps, this is the reason why photos of actors or furniture items are typically taken in diagonal views.

\section{Analysis of Content Coding}
\label{sec:coding}

Why did GVAE achieve better generalizable disentangling than MLVAE?  To gain insight into this question, we have conducted detailed analysis of content representation.  First of all, the most essential difference of MLVAE \cite{Bouchacourt:2018vt} from GVAE is the way inferring the content (equation~\ref{eq:gvae-inf}), where the posterior distribution is estimated as the product of Gaussians for individual instances (``evidence accumulation''):
\begin{align}
	q_\psi(z|x_1,\ldots,x_K) &= \frac{1}{Z}\prod_{k=1}^K\gaussd\left( h_\psi(x_k), s_\pi(x_k) \right) \label{eq:mlvae-inf-eviacc}
\end{align}
where $Z$ is the normalizing constant.  Since a product of Gaussians is a Gaussian, the above definition can be rewritten as:
\begin{align}
	\mbox{\eqref{eq:mlvae-inf-eviacc}} = 
	\gaussd\left( 
	  \frac{ \sum_k{h_\psi(x_k)}/s_\pi(x_k) }{ \sum_k{1/s_\pi(x_k)} },
	  \frac{ 1 }{ \sum_k 1/s_\pi(x_k) } 
	\right) \span\omit \label{eq:mlvae-inf} 
\end{align}
where all the multiplications and divisions are dimension-wise.  Note that the mean of the Gaussian has the form of weighted average where the weights are the precisions $1/s_\pi(x_k)$.  Although GVAE uses simple averaging (equation~\ref{eq:gvae-inf}), it is not a special case of MLVAE since the variance has a very different form.
We claim that this evidence accumulation technique unfortunately leads to view-dependent content representation in MLVAE unlike GVAE.

\begin{figure}[h]
\begin{center}
\includegraphics[width=8cm]{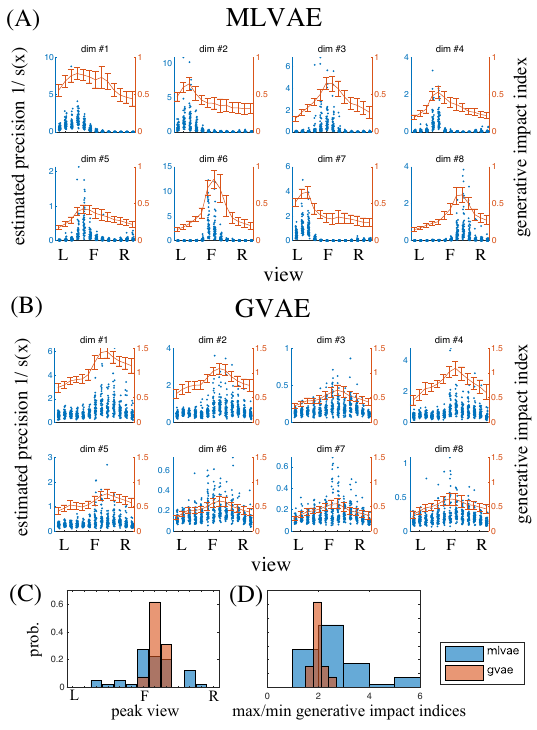}
\caption{Analysis of latent content representations.  (A) Blue: the distributions of estimated precisions $1/s(x)$ for each view for each of top 8 content dimensions in an MLVAE model (scale: $\times 10^3$).  Red: the view-wise normalized distances between two images generated from latent variables differing in a particular content dimension (generative impact index).  (B) Analogous results for a GVAE model.  (C) The distribution of peak views (in terms of generative impact indices) for the effective dimensions in the MLVAE or GVAE model (legend).  (D) The distribution of ratios of maximum and minimum generative impact indices in each model.   }
\label{fig:analysis}
\end{center}
\end{figure}

First of all, we found that, generally, MLVAE models had $\sim$3 times larger number of effective content dimensions than GVAE models.  To understand why, we picked up one example MLVAE model for Multi-PIE, which had 40 effective content dimensions, and inspected the structure of the content representation in the 8 most effective dimensions (i.e., with the largest s.d.'s).  In Figure~\ref{fig:analysis}A, each scatter plot (blue) shows the estimated precisions $1/s(x)$ for each view of test images (left-profile, frontal, right-profile, etc.), separately for each content dimension.  The estimated precisions tended to be peaked at a particular view while they went down to very low values elsewhere (e.g., dim \#1 had peak at 45$^\circ$ left profile), and the peak view was different from dimension to dimension.  This was starkly different in a GVAE model (which had 13 effective content dimensions).   Figure~\ref{fig:analysis}B shows the corresponding plot for this model, where the estimated precisions were only slightly higher for frontal views compared to profile views for most dimensions.  Recall that, in MLVAE, because of the weighted-average form (equation~\ref{eq:mlvae-inf}), each estimated mean is multiplied with the corresponding estimated precision.  Since the estimated precisions were very low for the non-peak views (Figure~\ref{fig:analysis}A), the content dimension would have strong ``impact'' only around the peak views.

To confirm this, we quantified how much each content dimension had impact on image generation.  We first estimated the latent variables $y=g(x)$ and $z=h(x)$ for each test image $x$ and then generated two new images using the estimated variables but modifying $d$-th content dimension.  Thereafter, we calculated the (normalized) Euclidean distance between the two generated images:
\[
I_d(x)=\frac{|| f(y,z^{-}_d) - f(y,z^{+}_d) ||}{(|| f(y,z^{-}_d) || + || f(y,z^{+}_d) ||)/2}
\]
where
\begin{align*}
z^{-}_d &= (z_1,\ldots,z_{d-1},\mu_d-3\sigma_d,z_{d+1},\ldots,z_M) \\
z^{+}_d &= (z_1,\ldots,z_{d-1},\mu_d+3\sigma_d,z_{d+1},\ldots,z_M)
\end{align*}
using the mean $\mu_d$ and s.d. $\sigma_d$ of $z_d$ (over the test data).
We called $I_d(x)$ ``generative impact index.''  If the two generated images are very different, then the content dimension has a large index; otherwise, it has a small index.  In Figure~\ref{fig:analysis}AB, each curve (red) shows generative impact indices for each view at each content dimension.  We can see that the generative impact consistently followed the magnitudes of precisions in both MLVAE and GVAE.  As a consequence, in MLVAE, each dimension had much more impact on image generation for high-precision views than low-precision views, whereas such contrast was more moderate in GVAE (Figure~\ref{fig:analysis}D).  Also, note that the peak views covered a broad range of views in MLVAE, while they were concentrated around the frontal view (Figure~\ref{fig:analysis}C).

Thus, MLVAE had view-dependent coding of content.  This in fact explains the generally larger number of effective content dimensionality in MLVAE.  That is, in MLVAE, different content dimensions seemed to be used for representing different views and therefore encoding all views required a larger number of dimensions, whereas each dimension seemed to play an equal role in all views in GVAE.

The view dependency of the estimated precisions in MLVAE may be related to uncertainty coding of facial features.  That is, since a single image provides only partial information for the content, the inference necessarily includes ambiguity.  For example, from a frontal face image, we are sure about the eyes, nose, and mouth, but less sure about the ears; from a profile face image, we are sure about the visible side of the face, but much less sure about the invisible side.  Thus, by taking uncertainty into account, we might be able to infer a more accurate content representation when integrating estimated content information from different views.  However, the view-dependent representation seems to go in the opposite direction to the goal of disentangling---to discover view-invariant representation.  This explains the reason for the observed lower performance of MLVAE in few-shot classification.

\section{Conclusion}
\label{sec:concl}

We have proposed group-based VAE as a novel method for learning disentangled representations from grouped data.  Despite the simplicity, our method achieved effective disentangling of content and transformation that generalized for test images with new contents.  We quantitatively evaluated generalization performance using few-shot classification and showed superiority of our approach to MLVAE for most datasets.  Our detailed analysis revealed that the performance difference was due to the evidence accumulation technique used in MLVAE causing transformation dependency of content representation.

Future investigation may pursue improvement of disentangling by incorporating some additional constraints such as adversarial learning.  Also, disentangling for more realistic and irregular dataset would be important.   Lastly, as the present model was initially inspired by findings in visual neuroscience and started as a continuation of our previous theoretical study on the primate visual system \cite{Hosoya:2016ct,Hosoya:2017bl}, we are keen to investigate how our model here could serve to explain the underlying learning principle in the higher vision.

\section*{Acknowledgments}

The author expresses special thanks to Aapo Hyv\"arinen, Mohammad Emtiyaz Khan, and Motoaki Kawanabe for precious comments and suggestions.  This work has been supported by the Commissioned Research of National Institute of Information and Communications Technology (1940201), the New Energy and Industrial Technology Development Organization (P15009), and Grants-in-Aid for Scientific Research (18H05021, 18K11517, and 19H04999). 

\appendix

\section{Dataset Details}
\label{sec:dataset-add}

\paragraph{Multi-PIE}

The original dataset \cite{Gross:2010fj} consists of natural face images of 337 subjects in 15 views, 3 expressions, and 4 (cloth/hair) fashions.  We used only images under a medium lighting (condition 9) and cropped and resized them using the manual landmark annotations in \cite{ElShafey:2013kw} ($64\times 64\times 3$ pixels).  We split the data into a training set with $\sim$32K images of 268 subjects and a test set with $\sim$6K images of the remaining 69 subjects.  We formed each group by randomly choosing face images of the same subject and fashion, but varying the view and the expression.  

\paragraph{Chairs} 

The original dataset \cite{Dosovitskiy:2015ux} consists of synthetic multi-viewed chair images of various types rendered from 3D models.  We used a convenient subset \cite{Yang:2016wg} including images ($64\times 64\times 3$ pixels) of 809 chair types in 62 views (31 horizontal and 2 vertical angles).  We split the data into a training set with $\sim$40K images of 650 chair types and a test set with $\sim$10K images of the remaining 159 chair types.  We formed each group by randomly selecting chair images of the same type but varying the view.  

\paragraph{KTH} 

The original dataset \cite{Schuldt:2004ee} consists of video clips of human motion of 6 types by 25 subjects, where each clip shows one type of motion by a single subject with varied settings (background, motion direction, cloth, camerawork, etc.).  From a subset with pedestrian motions (walking, running, and jogging), we created a training set with image frames ($64\times 64$ pixels) from 960 video clips of 20 subjects and a test set with image frames from 240 video clips of the remaining 5 subjects; we removed frames showing no person.   We formed each group by randomly selecting image frames in the same video clip.   

\paragraph{Sprites}

The original dataset \cite{Reed:2015uw} consists of synthetic game character images.  A structured label is given to each image to specify the features of the character; the images are varied for the pose and wear.  We assigned a unique id to each different structured label.  We split the data into a training set with $\sim$ 89K images of 500 ids and a test set with $\sim$ 19K images of 100 other ids.  We formed each group by randomly selecting images of the same id but varying the pose and wear.  

\paragraph{NORB}

The original dataset \cite{LeCun:2004if} consists of images of 50 toys with various views and lightings.  We split the data into training and test sets with disjoint ids (each $\sim$24k, 25 ids).  We formed each group by randomly selecting images of the same id but varying the view and lighting.

\section{Architecture Details}
\label{sec:architecture-add}

Each model used convolutional neural nets as the encoders ($g$, $r$, $h$, and $s$) and a deconvolutional neural net as the decoder ($f$).  Each encoder had three convolution layers with 64 filters (kernel $5\times 5$; stride $2$; padding 2) followed by two fully connected layers ($64$ intermediate and $2$ or $3$ output units for $g$ and $r$; $100$ intermediate  and $100$ units for $h$ and $s$).  The decoder $f$ had two fully connected layers ($102$ or $103$ output and $128$ intermediate units) followed by three transposed convolutional layers with 64 filters (kernel $6\times 6$; upsampling $2$; cropping $2$).  An RELU layer was inserted between convolutional or fully connected layers; the encoders $r$ and $s$ had an additional nonlinearity layer after the top layer to ensure positivity: $F(a)=\exp(a/2)$.  Only for an MLVAE model, we let the deep net $s$ encode the precision instead of the variance.

\pagebreak %
\bibliography{haruo.bib}

\bibliographystyle{named}

\end{document}

%% file: figs/nshot.txt
MultiPIE & {\bf 44.3}$\pm$3.2 & 24.5$\pm$3.9 & 9.1$\pm$0.8 & {\bf 64.0}$\pm$2.0 & 48.5$\pm$4.3 & 19.0$\pm$1.6 & 0.3\\
Chairs & {\bf 58.4}$\pm$6.7 & 53.3$\pm$5.2 & 18.4$\pm$3.0 & {\bf 82.7}$\pm$3.9 & 80.7$\pm$4.3 & 41.3$\pm$6.3 & 0.6\\
KTH & 27.1$\pm$1.6 & {\bf 30.1}$\pm$0.6 & 14.2$\pm$1.7 & 48.9$\pm$1.8 & {\bf 55.3}$\pm$0.8 & 35.6$\pm$3.1 & 0.4\\
Sprites & {\bf 81.3}$\pm$7.4 & 53.5$\pm$16.7 & 5.0$\pm$1.4 & {\bf 84.7}$\pm$4.8 & 65.0$\pm$14.6 & 9.5$\pm$3.1 & 1.0\\
NORB & {\bf 37.5}$\pm$1.3 & 31.2$\pm$1.9 & 12.9$\pm$3.7 & {\bf 44.7}$\pm$2.1 & 42.7$\pm$2.9 & 21.8$\pm$7.0 & 4.0\\

%% file: gendis-ijcai19.bbl
\begin{thebibliography}{}

\bibitem[\protect\citeauthoryear{Bouchacourt \bgroup \em et al.\egroup
  }{2018}]{Bouchacourt:2018vt}
Diane Bouchacourt, Ryota Tomioka, and Sebastian Nowozin.
\newblock {Multi-Level Variational Autoencoder: Learning Disentangled
  Representations from Grouped Observations}.
\newblock In {\em AAAI Conference on Artificial Intelligence}, 2018.

\bibitem[\protect\citeauthoryear{Chen \bgroup \em et al.\egroup
  }{2016}]{Chen:2016tp}
Xi~Chen, Yan Duan, Rein Houthooft, John Schulman, Ilya Sutskever, and Pieter
  Abbeel.
\newblock {InfoGAN - Interpretable Representation Learning by Information
  Maximizing Generative Adversarial Nets.}
\newblock {\em Advances in neural information processing systems}, 2016.

\bibitem[\protect\citeauthoryear{Chen \bgroup \em et al.\egroup
  }{2018}]{Chen:2017tk}
Micka{\"e}l Chen, Ludovic Denoyer, and Thierry Arti{\`e}res.
\newblock {Multi-View Data Generation Without View Supervision}.
\newblock In {\em International Conference on Learning Representations},
  November 2018.

\bibitem[\protect\citeauthoryear{Cheung \bgroup \em et al.\egroup
  }{2014}]{Cheung:2014ug}
Brian Cheung, Jesse~A Livezey, Arjun~K Bansal, and Bruno~A Olshausen.
\newblock {Discovering Hidden Factors of Variation in Deep Networks}.
\newblock In {\em International Conference on Learning Representations,
  Workshop}, December 2014.

\bibitem[\protect\citeauthoryear{Denton and Birodkar}{2017}]{Denton:2017uf}
Emily Denton and Vighnesh Birodkar.
\newblock {Unsupervised Learning of Disentangled Representations from Video}.
\newblock {\em Advances in neural information processing systems}, May 2017.

\bibitem[\protect\citeauthoryear{Dosovitskiy and
  Springenberg}{2015}]{Dosovitskiy:2015ux}
A~Dosovitskiy and J~Tobias Springenberg.
\newblock {Learning to generate chairs with convolutional neural networks}.
\newblock {\em Computer Vision and Pattern Recognition}, 2015.

\bibitem[\protect\citeauthoryear{El~Shafey \bgroup \em et al.\egroup
  }{2013}]{ElShafey:2013kw}
Laurent El~Shafey, Chris McCool, Roy Wallace, and S{\'e}bastien Marcel.
\newblock {A scalable formulation of probabilistic linear discriminant
  analysis: applied to face recognition.}
\newblock {\em IEEE Transactions on Pattern Analysis and Machine Intelligence},
  35(7):1788--1794, July 2013.

\bibitem[\protect\citeauthoryear{F{\"o}ldi{\'a}k}{1991}]{Foldiak:1991up}
P~F{\"o}ldi{\'a}k.
\newblock {Learning invariance from transformation sequences}.
\newblock {\em Neural Computation}, 3(2):194--200, 1991.

\bibitem[\protect\citeauthoryear{Goodfellow \bgroup \em et al.\egroup
  }{2014}]{Goodfellow:2014td}
I~Goodfellow, J~Pouget-Abadie, and M~Mirza.
\newblock {Generative adversarial nets}.
\newblock {\em Advances in neural information processing systems}, 2014.

\bibitem[\protect\citeauthoryear{Gross \bgroup \em et al.\egroup
  }{2010}]{Gross:2010fj}
Ralph Gross, Iain Matthews, Jeff Cohn, Takeo Kanade, and Simon Baker.
\newblock {Multi-PIE.}
\newblock {\em Proceedings of the International Conference on Automatic Face
  and Gesture Recognition. IEEE International Conference on Automatic Face {\&}
  Gesture Recognition}, 28(5):807--813, May 2010.

\bibitem[\protect\citeauthoryear{Higgins \bgroup \em et al.\egroup
  }{2016}]{Higgins:2016vm}
Irina Higgins, Loic Matthey, Arka Pal, Christopher Burgess, Xavier Glorot,
  Matthew Botvinick, Shakir Mohamed, and Alexander Lerchner.
\newblock {beta-VAE: Learning Basic Visual Concepts with a Constrained
  Variational Framework}.
\newblock In {\em International Conference on Learning Representations},
  November 2016.

\bibitem[\protect\citeauthoryear{Hosoya and
  Hyv{\"a}rinen}{2016}]{Hosoya:2016ct}
H~Hosoya and A~Hyv{\"a}rinen.
\newblock {Learning Visual Spatial Pooling by Strong PCA Dimension Reduction}.
\newblock {\em Neural Computation}, 28:1249--1263, 2016.

\bibitem[\protect\citeauthoryear{Hosoya and
  Hyv{\"a}rinen}{2017}]{Hosoya:2017bl}
Haruo Hosoya and Aapo Hyv{\"a}rinen.
\newblock {A mixture of sparse coding models explaining properties of face
  neurons related to holistic and parts-based processing}.
\newblock {\em PLoS Computational Biology}, 13(7):e1005667, July 2017.

\bibitem[\protect\citeauthoryear{Kingma and Ba}{2015}]{Kingma:2014us}
D~Kingma and J~Ba.
\newblock {Adam: A method for stochastic optimization}.
\newblock In {\em International Conference on Learning Representations}, 2015.

\bibitem[\protect\citeauthoryear{Kingma and Welling}{2014}]{Kingma:2013tz}
D~P Kingma and M~Welling.
\newblock {Auto-encoding variational bayes}.
\newblock In {\em International Conference on Learning Representations}, 2014.

\bibitem[\protect\citeauthoryear{Kingma \bgroup \em et al.\egroup
  }{2014}]{Kingma:2014uq}
D~P Kingma, S~Mohamed, and D~J Rezende.
\newblock {Semi-supervised learning with deep generative models}.
\newblock {\em Advances in neural information processing systems}, 2014.

\bibitem[\protect\citeauthoryear{Lample \bgroup \em et al.\egroup
  }{2017}]{Lample:2017vt}
Guillaume Lample, Neil Zeghidour, Nicolas Usunier, Antoine Bordes, Ludovic
  Denoyer, and Marc'Aurelio Ranzato.
\newblock {Fader Networks:Manipulating Images by Sliding Attributes}.
\newblock {\em Advances in neural information processing systems}, pages
  5967--5976, 2017.

\bibitem[\protect\citeauthoryear{LeCun \bgroup \em et al.\egroup
  }{2004}]{LeCun:2004if}
Yann LeCun, Fu~Jie Huang, and L{\'e}on Bottou.
\newblock {Learning Methods for Generic Object Recognition with Invariance to
  Pose and Lighting.}
\newblock {\em Computer Vision and Pattern Recognition}, 2004.

\bibitem[\protect\citeauthoryear{Mathieu \bgroup \em et al.\egroup
  }{2016}]{Mathieu:2016tn}
Micha{\"e}l Mathieu, Junbo~Jake Zhao, Pablo Sprechmann, Aditya Ramesh, and Yann
  LeCun.
\newblock {Disentangling factors of variation in deep representation using
  adversarial training.}
\newblock {\em Advances in neural information processing systems}, 2016.

\bibitem[\protect\citeauthoryear{Reed \bgroup \em et al.\egroup
  }{2015}]{Reed:2015uw}
Scott~E Reed, Yi~Zhang, Yuting Zhang, and Honglak Lee.
\newblock {Deep Visual Analogy-Making.}
\newblock {\em Advances in neural information processing systems}, 2015.

\bibitem[\protect\citeauthoryear{Sch{\"u}ldt \bgroup \em et al.\egroup
  }{2004}]{Schuldt:2004ee}
Christian Sch{\"u}ldt, Ivan Laptev, and Barbara Caputo.
\newblock {Recognizing Human Actions - A Local SVM Approach.}
\newblock In {\em International Conference on Pattern Recognition}, pages
  32--36 Vol.3. IEEE, 2004.

\bibitem[\protect\citeauthoryear{Siddharth \bgroup \em et al.\egroup
  }{2017}]{Siddharth:2017ud}
N~Siddharth, Brooks Paige, Jan-Willem van~de Meent, Alban Desmaison, Frank
  Wood, Noah~D Goodman, Pushmeet Kohli, and Philip H~S Torr.
\newblock {Learning Disentangled Representations with Semi-Supervised Deep
  Generative Models.}
\newblock {\em Advances in neural information processing systems}, 2017.

\bibitem[\protect\citeauthoryear{Tenenbaum and
  Freeman}{2000}]{Tenenbaum:2000wj}
J~B Tenenbaum and W~T Freeman.
\newblock {Separating style and content with bilinear models.}
\newblock {\em Neural Computation}, 12(6):1247--1283, June 2000.

\bibitem[\protect\citeauthoryear{Wang \bgroup \em et al.\egroup
  }{2017}]{Wang:2017taa}
Chaoyue Wang, Chaohui Wang, Chang Xu, and Dacheng Tao.
\newblock {Tag Disentangled Generative Adversarial Networks for Object Image
  Re-rendering}.
\newblock {\em International Joint Conference on Artificial Intelligence},
  August 2017.

\bibitem[\protect\citeauthoryear{Yang \bgroup \em et al.\egroup
  }{2015}]{Yang:2016wg}
Jimei Yang, Scott Reed, Ming-Hsuan Yang, and Honglak Lee.
\newblock {Weakly-supervised Disentangling with Recurrent Transformations for
  3D View Synthesis}.
\newblock {\em Advances in neural information processing systems}, 2015.

\end{thebibliography}
